\documentclass[11pt]{article}

\usepackage{times}
\usepackage{graphicx}
\usepackage{amsmath, amssymb}
\usepackage{hyperref}
\usepackage{geometry}
\title{Representation Stability in a Minimal Continual Learning Agent}

\author{
Vishnu Subramanian \\
AI/ML Computational Science Associate Manager \\
Accenture \\
\texttt{vishnu.subramanian93@gmail.com}
}

\date{}

\begin{document}
\maketitle

\begin{abstract}
Continual learning systems are increasingly deployed in environments where retraining or reset is infeasible, yet many approaches emphasize task performance rather than the evolution of internal representations over time. In this work, we study a minimal continual learning agent designed to isolate representational dynamics from architectural complexity and optimization objectives. The agent maintains a persistent state vector across executions and incrementally updates it as new textual data is introduced.

We quantify representational change using cosine similarity between successive normalized state vectors and define a stability metric over time intervals. Longitudinal experiments across eight executions reveal a transition from an initial plastic regime to a stable representational regime under consistent input. A deliberately introduced semantic perturbation produces a bounded decrease in similarity (from 0.9868 to 0.8957), followed by recovery and re-stabilization under subsequent coherent input.

These results demonstrate that meaningful stability–plasticity trade-offs can emerge in a minimal, stateful learning system without explicit regularization, replay, or architectural complexity. The work establishes a transparent empirical baseline for studying representational accumulation and adaptation in continual learning systems.
\end{abstract}

\section{Introduction}

Artificial learning systems are now being increasingly deployed and operated in real-world environments where retraining, reset and external supervision are infeasible. Despite this, many systems described as “continual learners” lack persistent internal state and / or meaningful measures of representational stability over time. As a result, learning is often conflated with short-term adaptation rather than long-term knowledge accumulation.

In this work, we study the behavior of a minimal continual learning agent that maintains an explicit internal representation across runs, and is exposed to a growing corpus of experience over time. Continual learning and catastrophic forgetting have been extensively studied in neural systems~\cite{parisi2019continual,french1999catastrophic}. Rather than trying to optimize task performance, our focus is on measuring how internal representations evolve, stabilize and adapt, as new information is encountered.

We argue that representation - not optimization - is the primary bottleneck in early-stage continual learning. To support this claim, we introduce simple, interpretable metrics based on normalized state vectors and cosine similarity, allowing us to quantify representational drift and stability to some extent across time. The role of representation learning as a foundational component of intelligent systems has been emphasized in prior work~\cite{bengio2013representation}.

Through longitudinal experiments, we observe a rapid transition from a highly plastic regime to a stable representational regime, followed by controlled adaptation when novel but related information is introduced. Notably, these dynamics emerge without any explicit regularization, forgetting mechanisms or architectural complexity.

These findings suggest that even minimal learning systems, when designed with persistent state and explicit self-comparison, can exhibit meaningful stability-plasticity trade-offs over time. By isolating representation dynamics from task performance and architectural complexity, this work establishes an empirical baseline for studying how learning systems accumulate, stabilize and adapt internal structure.

This baseline provides a foundation for future investigations into representation compression and efficiency (as emphasized in quantitative finance and production systems), stability-plasticity trade-offs observed in biological and neural systems, feedback-driven learning in the robotic and control domains, and learning under irreversible, delayed or energy-constrained conditions characteristic of engineering physics and space systems. At a more abstract level, the framework also invites exploration of learning as state evolution in high-dimensional spaces, drawing conceptual parallels with dynamical systems and quantum state representations. Finally, the simplicity of the present agent enables systematic study of arithmetic- and cognition-inspired representation strategies, such as those derived from abacus-based computation and symbolic methods such as Vedic mathematics, where structured compression and embodied representation play a central role in efficient human learning.

\section{Methods}
\subsection{Minimal Continual Learning Agent}

In this work, we study a minimal continual learning agent designed to isolate and expose the dynamics of internal representation over time. The agent operates in a repeated execution loop, processing a growing corpus of experience while maintaining a persistent internal state across runs. Crucially, the agent is not trained for task performance, optimization or reward maximization. Instead it is constructed as a measurement instrument for observing how representations accumulate, stabilize and adapt under continual exposure to new information.

At each execution, the agent ingests newly available textual data and updates an internal state that summarizes its accumulated experience. This internal state is explicitly stored and carried forward between runs, ensuring that learning is cumulative and irreversible. There is no mechanism for reset, replay or retraining from scratch. As a result, the agent's behavior more closely resembles an always-on system operating in an open-ended environment than a conventional machine learning model trained in discrete epochs.

The agent intentionally avoids architectural complexity. It does not employ deep neural networks, gradient-based optimization, replay buffers, or explicit forgetting mechanisms. This minimal design is deliberate: by stripping away task objectives and model sophistication, we aim to focus exclusively on representational change itself, rather than confounding it with performance-driven adaptation or algorithmic heuristics.

The agent is executed periodically (e.g. via a scheduled job), allowing observations of representation evolution across real time rather than within a single training session. This longitudinal execution model enables direct measurement of how internal state evolves from one run to the next as new information is incrementally incorporated. In this sense, the agent functions as a continual learner by construction, not by claim.

By framing the agent as a persistent, stateful process rather than a train-once model, this work shifts attention from task-centric evaluation to the dynamics of learning over time. The simplicity of the agent allows representational behavior to be observed transparently, providing a baseline against which more complex continual learning systems may later be compared.

\subsection{Learning Loop}

The agent operates through a simple, repeated learning loop that executes at each scheduled run. This loop defines how new experience is incorporated into the agent’s internal state and is designed to persist indefinitely without reset or episodic boundaries. The loop consists of five stages: observation, representation, update, comparison, and storage.

At each execution, the agent observes the current corpus of available textual data. The corpus is cumulative: documents introduced in earlier runs remain available in subsequent runs, ensuring that experience is not discarded or replayed selectively. This design choice enforces irreversibility, reflecting the constraints of always-on systems operating in open environments.

The observed text is transformed into a representational form by aggregating token occurrences into a fixed-dimensional vector. This representation summarizes the agent’s accumulated exposure up to the current time step, and serves as the agent's internal state. The representation is updated incrementally as new documents are added, rather than recomputing from scratch, reinforcing the notion of continual accumulation.

To quantify learning dynamics over time, the agent explicitly compares its current internal state with the state from the previous run. This self-comparison enables measurement of representational change without reference to external labels, tasks or performance objectives. The comparison step is central to the present study, as it allows representational stability and drift to be assessed longitudinally.

Finally, the updated internal state is stored persistently, and carried forward to the next execution. This stored state forms the basis for future updates and comparisons, closing the learning loop. Because the agent's state is preserved across runs, learning is defined not by convergence within a single session, but by the trajectory of internal representations across time.

This loop runs repeatedly as long as the agent remains active, with each iteration incorporating additional experience. By construction, the agent cannot “finish” learning or return to an earlier state. Instead, learning is framed as an ongoing process of accumulation, stabilization, and adaptation, observable through changes in the agent’s internal representation across successive executions.

\subsection{Representation and State Definition}

Vector-space representations have long been used in information retrieval and text modeling~\cite{manning2008introduction}. The agent’s internal state is represented as a fixed-dimensional numerical vector that encodes its accumulated exposure to textual experience. Each dimension of the vector corresponds to a unique token observed in the corpus, and the value along that dimension reflects the token’s aggregated frequency across all documents encountered up to the current run. This vector serves as the sole internal representation maintained by the agent.

Formally, let 
$\mathbf{s}_t \in \mathbb{R}^d$ 
denote the agent's internal state after execution t, where d is the size of the observed vocabulary at that time. As new documents are introduced, token counts are incrementally updated, expanding the dimensionality of the state vector when previously unseen tokens are encountered. The representation thus grows monotonically as experience accumulates, reflecting the open-ended nature of the learning environment.

To ensure comparability across runs, the state vector is normalized after each update. Normalization removes the influence of absolute document volume and allows representational change to be interpreted in terms of relative structure rather than scale. Without normalization, growth in corpus size would trivially dominate any measure of change, obscuring meaningful dynamics in representation evolution.

This representation is intentionally simple and fully interpretable. Unlike learned embeddings or latent neural activations, each dimension of the state vector corresponds directly to an observable property of the agent’s experience. This transparency allows representational change to be measured and reasoned about directly, without requiring auxiliary models or decoding procedures.

Importantly, the present work does not claim that token-frequency vectors are optimal representations for learning or generalization. Rather, the goal is to establish a minimal, explicit representation that makes internal state evolution measurable over time. By fixing this representational form and avoiding adaptive feature learning, the study isolates the dynamics of accumulation, stabilization, and drift in representation itself.

In this framework, learning is defined as a change in the agent’s internal state vector over successive executions. Stability corresponds to convergence in the direction of the normalized state vector, while adaptation manifests as controlled deviation in response to novel input. This definition allows learning behavior to be studied independently of task performance, optimization objectives, or external supervision.

\subsection{Representational Similarity Metric}

To quantify changes in the agent's internal representation over time, we employ cosine similarity between successive normalized state vectors. Cosine similarity measures the angular alignment between two vectors, capturing changes in representational structure independently of overall magnitude.

Given normalized internal states $\mathbf{s}_{t-1}$ and $\mathbf{s}_t$ at consecutive executions, the representational similarity is defined as:

\[\text{sim}(\mathbf{s}_{t-1}, \mathbf{s}_t) = \frac{\mathbf{s}_{t-1} \cdot \mathbf{s}_t}{\|\mathbf{s}_{t-1}\| \, \|\mathbf{s}_t\|}\]

Because state vectors are normalized, this measure reflects changes in the relative distribution of representational components rather than changes in corpus size or token volume. A similarity value close to 1 indicates that the direction of the internal representation has remained largely unchanged, while lower values indicate structural adaptation in response to new experience.

Cosine similarity is particularly well-suited for longitudinal analysis of continual learning systems, as it provides a stable, scale-invariant measure of representational drift across time. Unlike absolute distance metrics, cosine similarity does not grow trivially with accumulated experience, allowing meaningful comparisons even as the representation expands.

In the present study, representational similarity between successive runs serves as the primary observable for characterizing learning dynamics. High similarity values correspond to representational stability, while temporary decreases signal periods of increased plasticity or adaptation. Importantly, this metric does not assess correctness, performance, or task success; it captures only the degree of internal change induced by newly encountered information.

By focusing on self-similarity rather than external benchmarks, this approach enables learning behavior to be studied without reliance on labeled data, reward functions, or task-specific information. Representational similarity thus functions as a minimal, interpretable proxy for tracking how a continual learner’s internal state evolves over time.

\textbf{Representational Stability:}

We define representational stability over an interval $[t_1, t_2]$ as the persistence of directional alignment of the normalized state vector:

\[
S(t_1, t_2) = \frac{1}{t_2 - t_1} \sum_{t = t_1 + 1}^{t_2} \cos(\mathbf{s}_{t-1}, \mathbf{s}_t)
\]

Higher values of S indicate convergence toward a stable representational regime, while transient decreases correspond to periods of adaptation or plasticity.

\section{Results}

\subsection{Longitudinal Representation Dynamics}

\begin{figure}[t]
    \centering
    \includegraphics[width=0.9\linewidth]{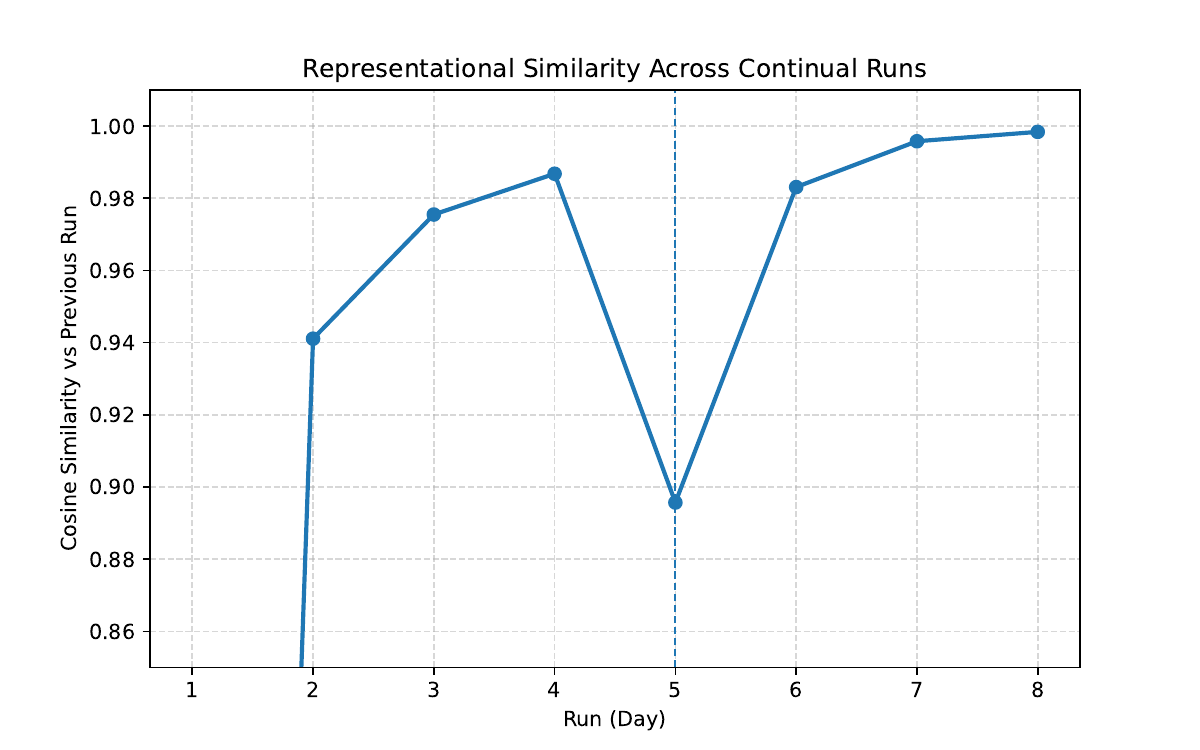}
    \caption{Cosine similarity between successive normalized internal state vectors across eight continual executions. Early runs exhibit rapid representational change, followed by stabilization under consistent input. A semantically orthogonal perturbation at Day 5 induces a bounded decrease in similarity to 0.8957, after which the representation recovers and re-stabilizes under subsequent coherent input.}
    \label{fig:similarity}
\end{figure}

Figure~\ref{fig:similarity} illustrates the evolution of representational similarity across successive runs.

We analyze the evolution of the agent’s internal representation across eight consecutive executions, each corresponding to the incremental introduction of new textual material. Representational change is quantified using cosine similarity between successive normalized state vectors, as defined in Section 2.4. This longitudinal view reveals distinct phases of learning behavior characterized by plasticity, stabilization, perturbation, and recovery.

During the initial executions (Days 1-3), the agent exhibits a plastic learning regime. Representational similarity increases rapidly across successive runs (from 0.0 at initialization to approximately 0.94 and 0.98), indicating substantial directional change as the internal state is formed and aligned in response to early exposure. In this phase, new information significantly reshapes the representation, reflecting the absence of an established internal structure.

By Day 4, the agent transitions into a stable regime. Representational similarity rises to approximately 0.99, suggesting that additional input produces only minor directional adjustments. At this stage, the internal representation exhibits strong coherence, and learning proceeds primarily through refinement rather than restructuring. This stabilization emerges without any explicit regularization, replay mechanism, or optimization objective.

At Day 5, a deliberate perturbation is introduced in the form of a semantically orthogonal document with largely disjoint vocabulary. This results in a pronounced decrease in representational similarity to approximately 0.90, making a transition back into a plastic adaptation regime. Importantly, the similarity drop remains bounded, indicating controlled adaptation rather than representational collapse or catastrophic forgetting. The agent incorporates novel information while preserving substantial alignment with its prior state.

Following the perturbation, the agent enters a recovery phase (days 6-8). When subsequent inputs return to the original semantic regime, representational similarity increases steadily, rising above 0.98 and ultimately returning to approximately 0.998 by Day 8. This re-stabilization demonstrates that representational stability is not permanently disrupted by novelty, but can be dynamically re-established through continued exposure to consistent input.

Together, these results demonstrate that even a minimal continual learning agent exhibits structured learning dynamics over time. The internal representation evolves rapidly during early exposure, stabilizes under consistent input, adapts in response to semantic novelty, and subsequently re-stabilizes when coherence is restored. These behaviors emerge solely from persistent state, incremental accumulation, and self-comparison, without reliance on task objectives, reward signals, or architectural complexity.

Table~\ref{tab:similarity} summarizes representational similarity, token accumulation, and vocabulary growth across all runs.

\begin{table}[t]
\centering
\begin{tabular}{c c c c}
\hline
Run (Day) & Tokens Seen & Vocabulary Size & Similarity vs Previous \\
\hline
1 & 48  & 44  & 0.0000 \\
2 & 86  & 73  & 0.9411 \\
3 & 119 & 92  & 0.9755 \\
4 & 153 & 111 & 0.9868 \\
5 & 393 & 273 & 0.8957 \\
6 & 430 & 292 & 0.9831 \\
7 & 464 & 301 & 0.9958 \\
8 & 497 & 312 & 0.9984 \\
\hline
\end{tabular}
\caption{Representational similarity, token accumulation, and vocabulary growth across successive continual runs.}
\label{tab:similarity}
\end{table}

\subsection{Response to Semantic Perturbation}

To probe the adaptability of the agent's internal representation, we introduce a deliberate semantic perturbation at Day 5. The perturbation consists of a single, extended document drawn from a markedly different conceptual and lexical domain than the preceding inputs, with minimal overlap in vocabulary. This intervention is designed to test whether the agent's representation can accommodate novel information without destabilizing previously accumulated structure.

The perturbation produces a clear and immediate effect on representational similarity. Cosine Similarity between the Day 4 and Day 5 internal states drops from approximately 0.99 to 0.8957, indicating a substantial directional shift in the normalized state vector. This change reflects genuine representational adaptation rather than trivial scaling effects, as magnitude differences are controlled through normalization.

Importantly, the observed deviation remains bounded. Despite the semantic novelty and vocabulary expansion introduced by the perturbation, the agent retains strong alignment with its prior representational state. This suggests that new information is assimilated into the existing structure rather than overwriting it. The absence of a more extreme similarity collapse indicates that the agent does not exhibit catastrophic forgetting under this form of novelty.

The bounded nature of the perturbation highlights a key property of the agent’s learning dynamics: plasticity is engaged selectively in response to semantic divergence, while representational coherence is preserved. This behavior emerges without explicit mechanisms for novelty detection, regularization, or memory protection. Instead, it arises naturally from the interaction between persistent state, incremental accumulation, and representational self-comparison.

These results demonstrate that even in a continual learning system, representational stability and plasticity are not mutually exclusive. Rather, they coexist as complementary modes of behavior, with the system transitioning smoothly between them as environmental input shifts. The semantic perturbation thus serves as a controlled probe of adaptive capacity, revealing that meaningful learning dynamics can emerge in the absence of task-driven optimization or architectural complexity.

\subsection{Vocabulary Growth and State Expansion}

In addition to representational similarity, we track auxiliary statistics related to the growth of the agent’s internal state, including cumulative token count and vocabulary size. These quantities provide a complementary view of learning dynamics, capturing the irreversible accumulation of experience underlying representational change.

Across all executions, both tokens observed and vocabulary size increase monotonically. Early runs exhibit rapid vocabulary expansion as the agent encounters new terms for the first time. As exposure continues within a consistent semantic regime, vocabulary growth slows, reflecting increasing reuse of previously observed tokens. This pattern aligns with the transition from a highly plastic regime to a stable representational regime observed in similarity metrics.

The semantic perturbation introduced at Day 5 produces a pronounced increase in vocabulary size, corresponding to the introduction of a large number of previously unseen tokens. Notably, this expansion occurs simultaneously with a decrease in representational similarity, indicating that the perturbation affects both the breadth of the representation and its directional alignment. However, vocabulary growth alone does not account for the observed similarity dynamics: subsequent recovery runs show continued, albeit slower, vocabulary expansion while representational similarity increases toward prior levels.

This decoupling highlights an important distinction between representational capacity and representational orientation. While the internal state continues to grow in dimensionality as new tokens are encountered, the normalized direction of the state vector can nevertheless stabilize. Learning, in this sense, is not characterized by cessation of growth, but by the convergence of relative structure within an expanding space.

Together with similarity-based analysis, vocabulary and token statistics reinforce the interpretation of learning as cumulative and irreversible. The agent does not discard or compress past experience; instead, it incorporates new information by expanding its internal state while maintaining coherence in representational direction. These properties emerge naturally from persistent state and incremental accumulation, without any explicit mechanisms for memory management or feature selection.

\section{Discussion}

This work examines representation dynamics in a minimal continual learning agent, deliberately decoupled from task objectives, reward signals, and architectural complexity. By focusing exclusively on persistent internal state and longitudinal self-comparison, the study isolates learning as a process of state evolution over time rather than performance optimization. The resulting dynamics - plasticity, stabilization, controlled adaptation, and recovery - emerge naturally from incremental accumulation and normalization, without explicit mechanisms designed to enforce them.

A central observation is that representational stability does not imply stagnation. Even as the agent’s internal state continues to grow in dimensionality and incorporate new information, the normalized direction of the representation converges under consistent input. This distinction between representational capacity and representational orientation clarifies how learning systems can remain adaptive while preserving coherence. Stability, in this sense, is not the absence of change but the convergence of relative structure within an expanding space.

The semantic perturbation experiment further demonstrates that adaptation can be both selective and reversible. Exposure to a semantically orthogonal document induces a measurable but bounded deviation in representational alignment, followed by re-stabilization when subsequent input returns to a familiar regime. This behavior suggests that stability-plasticity trade-offs can arise as an emergent property of persistent state and incremental accumulation, rather than requiring explicit regularization or forgetting mechanisms. The stability–plasticity dilemma has been explored in cognitive and neural models such as Adaptive Resonance Theory~\cite{grossberg1987art}.

These findings have implications for how continual learning systems are evaluated. Much of the existing literature emphasizes task performance, benchmark retention, or catastrophic forgetting under supervised objectives. In contrast, the present work proposes representation dynamics themselves as a first-class object of study. Measuring how internal structure evolves, stabilizes, and adapts over time provides insight into learning behavior that is orthogonal to task success, and may serve as a foundation for more complex systems.

Importantly, the minimal nature of the agent is a feature rather than a limitation. By removing confounding factors, the study reveals that meaningful learning dynamics can arise even in the absence of optimization, deep architectures, or reward-driven updates. This suggests that some properties often attributed to sophisticated learning algorithms may instead be rooted in more fundamental principles of accumulation, normalization, and self-comparison.

\section{Limitations and Future Work}

The present study is intentionally limited in scope. The agent employs a simple token-frequency representation and operates exclusively on static textual input. No mechanisms for compression, abstraction, embodiment, or action are included. As a result, the findings should not be interpreted as claims about optimal representations or general intelligence, but rather as a baseline characterization of representation dynamics in a minimal continual learner.

Several important limitations follow directly from this design. The representation grows without bound, raising questions about long-term scalability, memory efficiency, and compression. The agent does not interact with an environment or influence its inputs, precluding the study of feedback-driven or embodied learning. Additionally, similarity metrics capture only coarse-grained structural alignment and do not reflect higher-order semantic organization.

These limitations also define a rich space for future work. From the mathematical sciences, future extensions may incorporate structured representations, dimensionality reduction, information-theoretic measures, and dynamical systems analysis to better characterize state evolution. Concepts from optimization, control theory, and statistical mechanics may provide deeper insight into convergence, stability, and phase transitions in learning dynamics.

Incorporating ideas from robotics and autonomous systems opens the possibility of studying learning under physical constraints, delayed feedback, partial observability, and real-time operation. Persistent state evolution in such settings would connect representation stability to control performance, safety, and robustness. Similarly, ideas from autonomous aerial vehicles and space systems introduce learning under severe resource, reliability, and irreversibility constraints.

Insights from neuroscience and brain-machine interfaces suggest avenues for studying embodied representations,  sensorimotor coupling, and co-adaptation between internal state and external action. Biological learning systems achieve remarkable stability and adaptability with limited resources, offering inspiration for compression modularity, and hierarchical organization.

At a more fundamental level, theoretical physics and quantum computing motivate viewing learning as state evolution in high-dimensional spaces governed by constraints and transformations. Analogies with Hamilton dynamics, energy landscapes, and quantum state representations may yield new abstractions for understanding learning trajectories and stability.

The long-term goal of this research program is to develop a controlled framework for recursively self-improving agentic intelligence. By grounding each step in measurable representation dynamics and incremental extensions, future work aims to explore how learning systems can improve not only their knowledge, but their own mechanisms of learning, while remaining interpretable, stable, and bounded.

\section{Conclusion}

This paper presents a minimal continual learning agent designed to expose the dynamics of internal representation over time. By maintaining persistent state and measuring representational self-similarity across executions, the study demonstrates that learning can be observed as a process of accumulation, stabilization, adaptation, and recovery, independent of task performance or optimization objectives.

Through longitudinal experiments and a controlled semantic perturbation, we show that even simple learning systems can exhibit meaningful stability-plasticity trade-offs. Representations stabilize under consistent input, adapt in response to novelty, and re-stabilize when coherence is restored. These behaviors emerge without explicit architectural mechanisms, suggesting that fundamental properties of learning may arise from persistent state and incremental exposure alone.

By framing learning as state evolution rather than task optimization, this work establishes a transparent baseline for studying continual learning dynamics. The simplicity of the agent enables precise measurement and interpretation, providing a foundation upon which increasingly complex, embodied, and adaptive learning systems can be built.

Ultimately, understanding how representations evolve over time is essential for the development of reliable, long-lived learning agents. This work takes a first step toward that goal by demonstrating that stability and adaptability need not be engineered explicitly, but can emerge from minimal principles when learning is allowed to persist.

\bibliographystyle{plain}
\bibliography{references}

\end{document}